\pdfoutput=1

\documentclass[11pt]{article}

\usepackage[]{naacl2021}
\usepackage{times}
\usepackage{latexsym}
\usepackage{amsmath}
\usepackage{graphicx}
\usepackage{pgfplots}
\pgfplotsset{compat=1.16}
\usepackage{tabularx}
\usepackage{xcolor}
\usepackage{soul}
\usepackage{booktabs}
\usepackage{amssymb}
\usepackage{pifont}
\newcommand{\cmark}{\ding{51}}%
\newcommand{\xmark}{\ding{55}}%
\usepackage{changes}
\usepackage[framemethod=tikz]{mdframed}

\usepackage{times}
\usepackage{latexsym}

\usepackage[T1]{fontenc}

\usepackage[utf8]{inputenc}

\usepackage{microtype}

\title{Augmented SBERT: Data Augmentation Method for Improving Bi-Encoders for Pairwise Sentence Scoring Tasks}

\author{Nandan Thakur, Nils Reimers, Johannes Daxenberger and Iryna Gurevych \\
  Ubiquitous Knowledge Processing Lab (UKP-TUDA) \\
  Department of Computer Science, Technische Universit{\"a}t Darmstadt \\
  \href{https://www.informatik.tu-darmstadt.de/ukp/ukp_home/index.en.jsp}{\texttt{www.ukp.tu-darmstadt.de}}}

\begin{document}
\maketitle

\begin{abstract}
There are two approaches for pairwise sentence scoring: \textit{Cross-encoders}, which perform full-attention over the input pair, and \textit{Bi-encoders}, which map each input independently to a dense vector space. 
While cross-encoders often achieve higher performance, they are too slow for many practical use cases. 
Bi-encoders, on the other hand, require substantial training data and fine-tuning over the target task to achieve competitive performance. 
We present a simple yet efficient data augmentation strategy called \textit{Augmented SBERT}, where we use the cross-encoder to label a larger set of input pairs to augment the training data for the bi-encoder. We show that, in this process, selecting the sentence pairs is non-trivial and crucial for the success of the method.
We evaluate our approach on multiple tasks (in-domain) as well as on a domain adaptation task. \textit{Augmented SBERT} achieves an improvement of up to 6 points for in-domain and of up to 37 points for domain adaptation tasks compared to the original bi-encoder performance.\footnote{Code available: \href{https://www.sbert.net}{www.sbert.net}}
\end{abstract}

\section{Introduction}
Pairwise sentence scoring tasks have wide applications in NLP. They can be used in information retrieval, question answering, duplicate question detection, or clustering. 
An approach that sets new state-of-the-art performance for many tasks including pairwise sentence scoring is BERT \cite{devlin2018bert}. 
Both sentences are passed to the network and attention is applied across all tokens of the inputs. 
This approach, where both sentences are simultaneously passed to the network, is called \textit{cross-encoder} \cite{Humeau2020Poly-encoders}. 

\begin{figure}[t]
\centering
\begin{center}
    \includegraphics[trim=70 0 50 0,clip,width=0.45\textwidth]{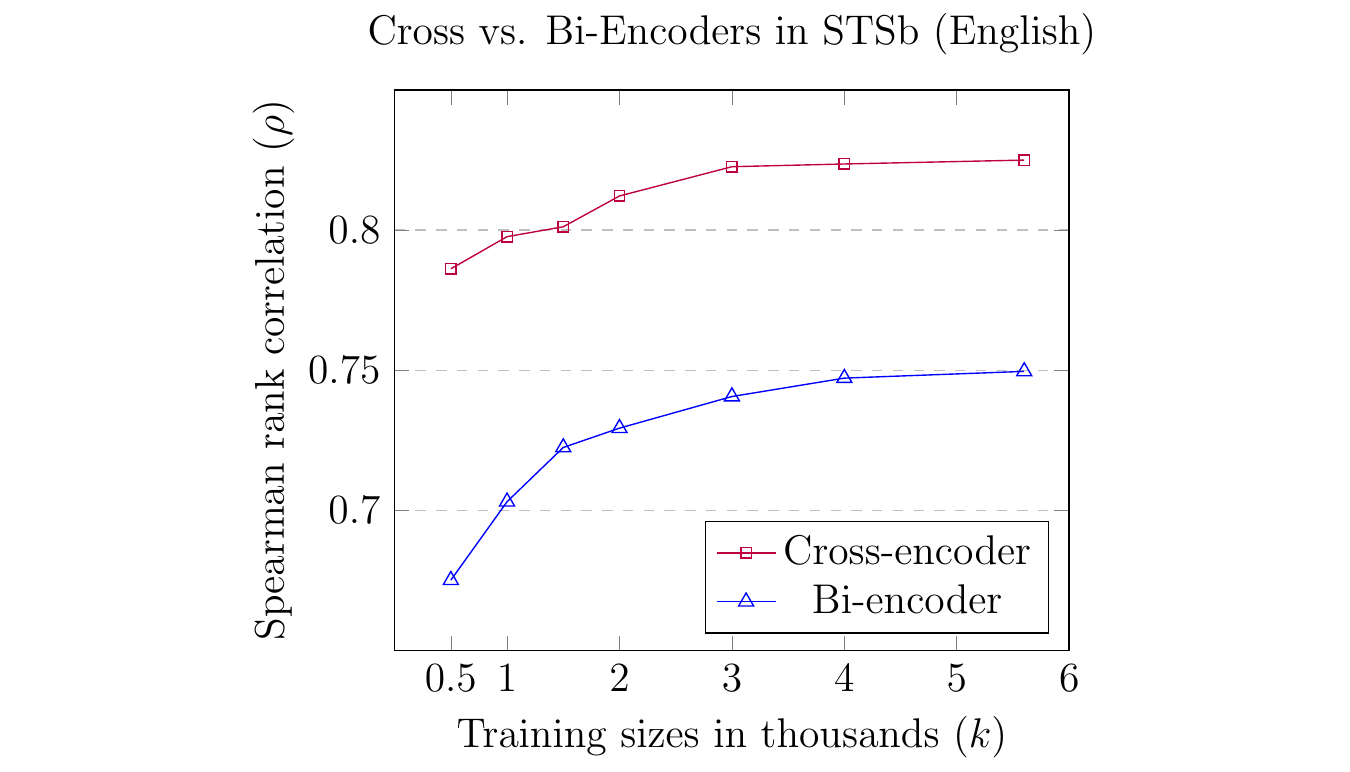}
    \captionof{figure}{Spearman rank correlation ($\rho$) test scores for different STS Benchmark (English) training sizes.}
    \label{fig:english-sts}
\end{center}
\end{figure}

A downside of cross-encoders is the extreme computational overhead for many tasks. For example, clustering of 10,000 sentences has a quadratic complexity with a cross-encoder and would require about 65 hours with BERT \cite{reimers-2019-sentence-bert}. End-to-end information retrieval is also not possible with cross-encoders, as they do not yield independent representations for the inputs that could be indexed. 
In contrast, \textit{bi-encoders} such as Sentence BERT (SBERT) \cite{reimers-2019-sentence-bert}  encode each sentence independently and map them to a dense vector space. 
This allows efficient indexing and comparison. For example, the complexity of clustering 10,000 sentences is reduced from 65 hours to about 5 seconds \cite{reimers-2019-sentence-bert}. Many real-world applications hence depend on the quality of bi-encoders.

A drawback of the SBERT bi-encoder is usually a lower performance in comparison with the BERT cross-encoder. 
We depict this in \autoref{fig:english-sts}, where we compare a fine-tuned cross-encoder (BERT) and a fine-tuned  bi-encoder (SBERT) over the popular English STS Benchmark dataset\footnote{\href{http://ixa2.si.ehu.es/stswiki/index.php/STSbenchmark}{http://ixa2.si.ehu.es/stswiki/index.php/STSbenchmark}} \cite{cer-etal-2017-semeval} for different training sizes and spearman rank correlation ($\rho$) on the test split.

This performance gap is the largest when little training data is available. 
The BERT cross-encoder can compare both inputs simultaneously, while the SBERT bi-encoder has to solve the much more challenging task of mapping inputs independently to a meaningful vector space which requires a sufficient amount of training examples for fine-tuning.
 
In this work, we present a data augmentation method, which we call \textit{Augmented SBERT} (AugSBERT), that uses a BERT cross-encoder to improve the performance for the SBERT bi-encoder. We use the cross-encoder to label new input pairs, which are added to the training set for the bi-encoder.  The SBERT bi-encoder is then fine-tuned on this larger augmented training set, which yields a significant performance increase. As we show, selecting the input pairs for soft-labeling with the cross-encoder is non-trivial and crucial for improving performance. 
Our method is easy to apply to many pair classification and regression problems, as we show in the exhaustive evaluation of our approach.

First, we evaluate the proposed AugSBERT method on four diverse tasks: Argument similarity, semantic textual similarity, duplicate question detection, and news paraphrase identification. We observe consistent performance increases of 1 to 6 percentage points over the state of the art SBERT bi-encoder's performance.
Next, we demonstrate the strength of AugSBERT in a domain adaptation scenario.
Since the bi-encoder is not able to map the new domain to a sensible vector space, the performance drop on the target domain for SBERT bi-encoders is much higher than for BERT cross-encoders. 
In this scenario, AugSBERT achieves a performance increase of up to 37 percentage points.

\section{Related Work}
Sentence embeddings are a well studied area in recent literature. Earlier techniques included unsupervised methods such as Skip-thought vectors \cite{10.5555/2969442.2969607} and supervised methods such as InferSent \cite{conneau-etal-2017-supervised} or USE \cite{cer-etal-2018-universal}. 
For pairwise scoring tasks, more recent sentence embedding techniques are also able to encode a pair of sentences jointly.
Among these, BERT \cite{devlin2018bert} can be used as a cross-encoder.
Both inputs are separated by a special \texttt{SEP} token and multi-head attention is applied over all input tokens. While the BERT cross-encoder achieves high performances for many sentence pair-tasks, a drawback is that no independent sentence representations are generated. This drawback was addressed by SBERT \cite{reimers-2019-sentence-bert}, which applies BERT independently on the inputs followed by mean pooling on the output to create fixed-sized sentence embeddings.

\newcite{Humeau2020Poly-encoders} showed that cross-encoders typically outperform bi-encoders on sentence scoring tasks. They proposed a third strategy (poly-encoders), that is in-between cross- and bi-encoders. Poly-encoders utilize two separate transformers, one for the candidate and one for the context. A given candidate is represented by one vector, while the context is jointly encoded with the candidates (similar to cross-encoders). Unlike cross-encoder's full self attention technique, poly-encoders apply attention between two inputs only at the top layer. Poly-encoders have the drawback that they are only practical for certain applications: The score function is not symmetric, i.e., they cannot be applied for tasks with a symmetric similarity relation. Further, poly-encoder representations cannot be efficiently indexed, causing issues for retrieval tasks with large corpora sizes. 

\newcite{chen-etal-2020-dipair} propose the DiPair architecture which, similar to our work, also uses a cross-encoder model to annotate unlabeled pairs for fine-tuning a bi-encoder model. 
DiPair focuses on inference speed and provides a detailed ablation for optimal bi-encoder architectures for performance versus speed trade-offs. The focus of our work are sampling techniques, which we find crucial for performance boosts in the bi-encoder model while keeping its architecture constant.

Our proposed data augmentation approach is based on semi-supervision \cite{10.1145/279943.279962} for in-domain tasks, which has been applied successfully for a wide range of tasks. \newcite{uva-etal-2018-injecting} train a SVM model with few gold samples and apply semi-supervision with pre-training neural networks. Another common strategy is to generate paraphrases of existent sentences, for example, by replacing words with synonyms \cite{wei-zou-2019-eda}, by using round-trip translation \cite{wei2018fast,xie2020unsupervised}, or with seq2seq-models \cite{kumar-etal-2019-submodular}. Other approaches generate synthetic data by using generative adversarial networks \cite{tanaka2019data}, by using a language model to replace certain words \cite{cbert-aug} or to generate complete sentences \cite{anaby2019not}. These data augmentation approaches have in common that they were applied to single sentence classification tasks. In our work, we focus on \textit{sentence pair tasks}, for which we need to generate suitable sentence pairs. As we show, randomly combining sentences is insufficient. Sampling appropriate pairs has a decisive impact on performance 
which corresponds to recent findings on similar datasets \cite{peinelt-etal-2019-aiming}.

\section{Methods}
In this section we present Augmented SBERT for diverse sentence pair in-domain tasks. We also evaluate our method for domain adaptation tasks.

\subsection{Augmented SBERT}

\begin{figure}[h]
    \centering
    \begin{center}
        \includegraphics[trim=75 5 60 0,clip,width=0.4\textwidth]{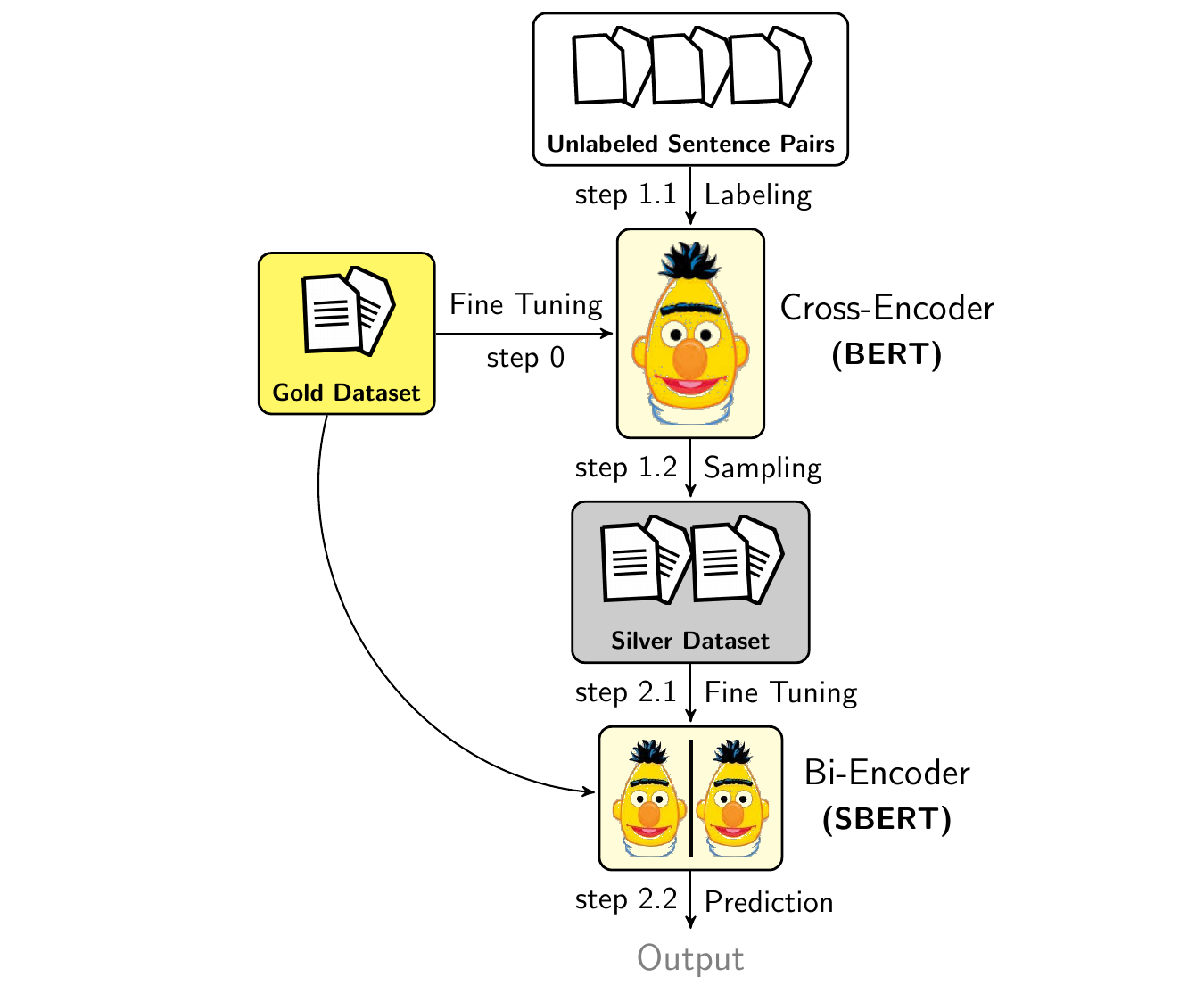}
        \captionof{figure}{Augmented SBERT In-domain approach}
        \label{fig:augsbert}
    \end{center}
    \end{figure}

     Given a pre-trained, well-performing cross-encoder, we sample sentence pairs according to a certain \textit{sampling strategy} (discussed later) and label these using the cross-encoder. We call these weakly labeled examples the \textit{silver dataset} and they will be merged with the gold training dataset. We then train the bi-encoder on this extended training dataset. We refer to this model as Augmented SBERT (AugSBERT). The process is illustrated in \autoref{fig:augsbert}.
    
    \paragraph{Pair Sampling Strategies} The novel sentence pairs, that are to be labeled with the cross-encoder, can either be new data or we can re-use individual sentences from the gold training set and re-combine pairs. In our in-domain experiments, we re-use the sentences from the gold training set. This is of course only possible if not all combinations have been annotated. However, this is seldom the case as there are $n\times(n-1)/2$ possible combinations for $n$ sentences. Weakly labeling all possible combinations would create an extreme computational overhead, and, as our experiments show, would likely not lead to a performance improvement. Instead, using the right sampling strategy is crucial to achieve a performance improvement.
    
    \textbf{\textit{Random Sampling (RS):}} We randomly sample a sentence pair and weakly label it with the cross-encoder. Randomly selecting two sentences usually leads to a dissimilar (negative) pair; positive pairs are extremely rare. This skews the label distribution of the silver dataset heavily towards negative pairs.
    
    \textbf{\textit{Kernel Density Estimation (KDE):}} We aim to get a similar label distribution for the silver dataset as for the gold training set. To do so, we weakly label a large set of randomly sampled pairs and then keep only certain pairs. For classification tasks, we keep all the positive pairs. Subsequently we randomly sample out negative pairs from the remaining dominant negative silver-pairs, in a ratio identical to the gold dataset training distribution (positives/negatives). For regression tasks, we use kernel density estimation (KDE) to estimate the continuous density functions $F_{gold}(s)$ and $F_{silver}(s)$ for scores $s$. 
    We try to minimize KL Divergence \cite{kullback1951} between distributions
 
    using a sampling function which retains a sample with score $s$ with probability $Q(s)$:

    \begin{small}
    \begin{equation*}
      Q(s) =
        \begin{cases}
            \hfil 1 & \text{if } F_{gold}(s) \geq F_{silver}(s) \\ 
          \\
          \dfrac{F_{gold}(s)}{F_{silver}(s)} & \text{if } F_{gold}(s) < F_{silver}(s) \\
        \end{cases}   
    \end{equation*}
    \end{small}
    
    Note, that the KDE sampling strategy is computationally inefficient as it requires labeling many, randomly drawn samples, which are later discarded.
    
    \textbf{\textit{BM25 Sampling (BM25):}} In information retrieval, the Okapi BM25 \cite{Amati2009} algorithm is based on lexical overlap and is commonly used as a scoring function by many search engines. We utilize ElasticSearch\footnote{\href{https://www.elastic.co/}{https://www.elastic.co/}} for the creation of indices which helps in fast retrieval of search query results. For our experiments, we index every unique sentence, query for each sentence and retrieve the top $k$ similar sentences. These pairs are then weakly labeled using the cross-encoder. Indexing and retrieving similar sentences is efficient and all weakly labeled pairs will be used in the silver dataset.
    
     \textbf{\textit{Semantic Search Sampling (SS):}} A drawback of BM25 is that only sentences with lexical overlap can be found. Synonymous sentences with no or little lexical overlap will not be returned, and hence, not be part of the silver dataset. We train a bi-encoder (SBERT) on the gold training set as described in \autoref{sec:experimental_setup} and use it to sample further, similar sentence pairs. We use cosine-similarity and retrieve for every sentence the top $k$ most similar sentences in our collection. For large collections, approximate nearest neighbour search like Faiss\footnote{\href{https://github.com/facebookresearch/faiss}{https://github.com/facebookresearch/faiss}} could be used to quickly retrieve the $k$ most similar sentences. 
     
    \textbf{\textit{BM25~+~Semantic Search Sampling (BM25-S.S.):}} We apply both BM25 and Semantic Search (S.S.) sampling techniques simultaneously. Aggregating the strategies helps capture the lexical and semantically similar sentences but skews the label distribution towards negative pairs.

    \paragraph{Seed Optimization} \newcite{dodge2020fine} show a high dependence on the random seed for transformer based models like BERT, as it converges to different minima that generalize differently to unseen data \cite{LeCun1998, Erhan2010, reimers-gurevych-2017-reporting}. This is especially the case for small training datasets. In our experiments, we apply \textit{seed optimization}: We train with 5 random seeds and select the model that performs best on the development set. In order to speed this up, we apply \textit{early stopping} at 20\% of the training steps and only continue training the best performing model until the end. We empirically found that we can predict the final score with high confidence at 20\% of the training steps (\autoref{sec:seed-opt}).

\begin{figure}[htb]
    \centering
    \begin{center}
        \includegraphics[trim=0 10 0  0,clip,width=0.5\textwidth]{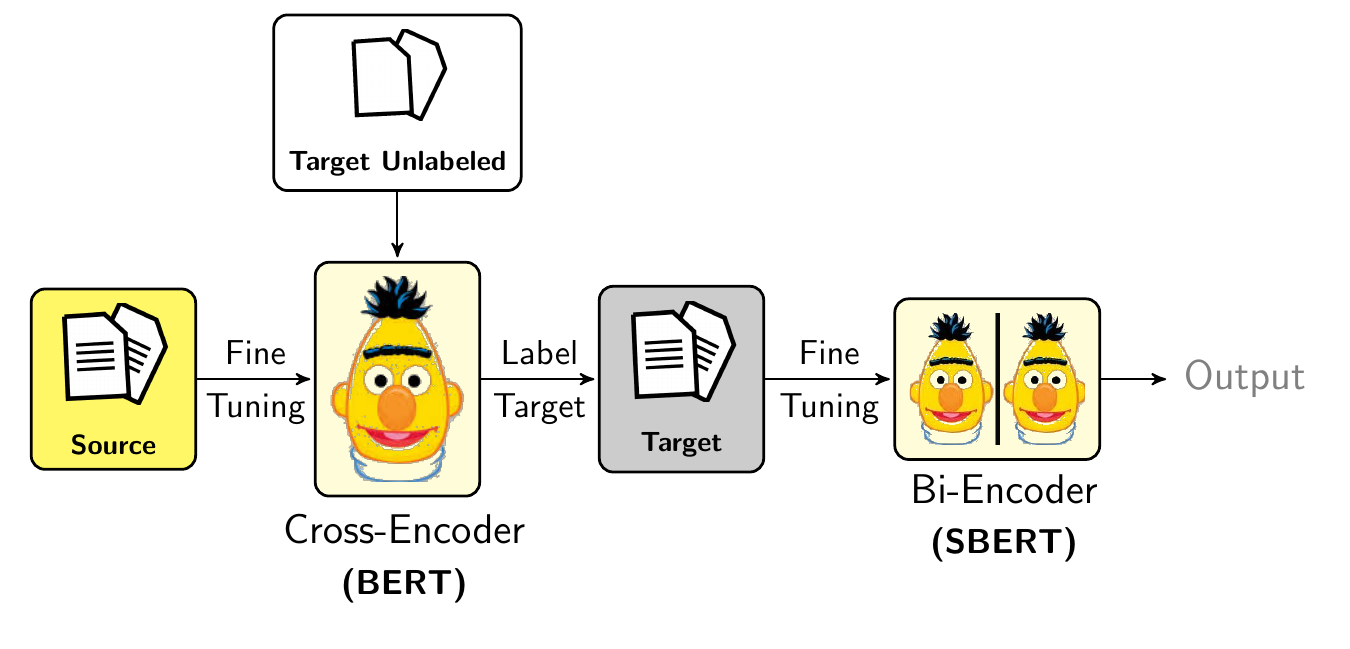}
        \captionof{figure}{Domain adaptation with AugSBERT.}
        \label{fig:augsbert-cross-domain}
    \end{center}
\end{figure}

\subsection{Domain Adaptation with AugSBERT}
\label{sec:DomainAdaptationwithAugSBERT}

Until now we discussed Augmented SBERT for in-domain setups, i.e., when the training and test data are from the same domain. 
However, we expect an even higher performance gap of SBERT on out-of-domain data.
This is because SBERT fails to map sentences with unseen terminology to a sensible vector space. Unfortunately, annotated data for new domains is rarely available. 

Hence, we evaluate the proposed data augmentation strategy for domain adaptation: We first fine-tune a cross-encoder (BERT) over the source domain containing pairwise annotations. After fine-tuning, we use this fine-tuned cross-encoder to label the target domain. Once labeling is complete, we train the bi-encoder (SBERT) over the labeled target domain sentence pairs (\autoref{fig:augsbert-cross-domain}).

\section{Datasets}

    \begin{table*}[t!]
        \centering
        \small
        \begin{tabular}{ c | c c c c c }
            \toprule
               \textbf{Dataset} & \textbf{Spanish-STS} & \textbf{BWS (cross-topic)}  & \textbf{BWS (in-topic)} & \textbf{Quora-QP} & \textbf{MRPC} \\
             \midrule
             \# training-samples & 1,400 & 2125 & 2471 & 10,000 & 4,340 \\
             \# development-samples & 220 & 425 & 478 & 3,000 & 731 \\
             \# testing-samples & 250 & 850 & 451 & 3,000 & 730 \\
             \midrule
             \# total-samples & \textbf{1,870} & \textbf{3,400} & \textbf{3,400} & \textbf{16,000} & \textbf{5,801} \\
             \bottomrule
        \end{tabular}
        \centering
        \caption{Summary of all datasets being used for diverse in-domain sentence pair tasks in this paper.}
        \label{tab:my_label_1}
    \end{table*}

\begin{table*}[t]
    \centering
    \small
        \begin{tabular}{l|p{5cm}p{7cm}|c}
        \toprule
        \textbf{Dataset} & \textbf{Sentence 1} & \textbf{Sentence 2} & \textbf{Score}\\ \midrule
        BWS & Cloning treats children as objects.
        & It encourages parents to regard their children as property. & 0.89 \\ \midrule
        
        Quora-QP & How does one cook broccoli? & What are the best ways to cook broccoli? & 1 \\ \midrule
        
        Spanish-STS & Dos hombres en trajes rojos practicando artes marciales.
        & Dos hombre en uniformes de artes marciales entrenando.
        & 0.80 \\ \midrule
        MRPC & The DVD-CCA then appealed to the state Supreme Court. 
        & DVD CCA appealed that decision to the U.S. Supreme Court. & 1 \\ \bottomrule
        \end{tabular}
        \caption{Dataset examples for our in-domain tasks. We report the normalized similarity score $[0,1]$ for regression tasks and the binary label $\{0,1\}$ for classification tasks.}
        \label{tab:dataset-examples}
    \end{table*}

Sentence pair scoring can be differentiated in regression and classification tasks. Regression tasks assign a score to indicate the similarity between the inputs. For classification tasks, we have distinct labels, for example, \textit{paraphrase} vs. \textit{non-paraphrase}. 

\subsection{Single-Domain Datasets}

In our single-domain (i.e. in-domain) experiments, we use two sentence pair regression tasks: semantic textual similarity and argument similarity. 
Furthermore, we use two binary sentence pair classification tasks: Duplicate question detection and news paraphrase identification. 
Examples for all datasets are given in \autoref{tab:dataset-examples}.

    \textbf{SemEval Spanish STS:} Semantic Textual Similarity (STS)\footnote{\href{https://ixa2.si.ehu.es/stswiki}{https://ixa2.si.ehu.es/stswiki}} is the task of assessing the degree of similarity between two sentences over a scale ranging from $[0,5]$ with $0$ indicating no semantic overlap and $5$ indicating identical content \cite{agirre2016semeval}. We choose Spanish STS data to test our methods for a different language than English. For our training and development dataset, we use the datasets provided by SemEval STS 2014 \cite{agirre-etal-2014-semeval} and SemEval STS 2015 \cite{agirre2015semeval}. These consist of  annotated sentence pairs from news articles and from Wikipedia. As test set, we use SemEval STS 2017 \cite{cer-etal-2017-semeval}, which annotated image caption pairs from SNLI \cite{snli:emnlp2015}. For all our experiments, we normalise the original similarity scores to $[0,1]$ by dividing the score by 5.
    
    \textbf{BWS Argument Similarity Dataset (BWS):} Existing similarity datasets have the disadvantage that the sentence pair selection/sampling process is not always comprehensible.
    To overcome this limitation, we create and publicly release a novel dataset\footnote{Public Data Release (BWS Argument Similarity Corpus): \href{https://tudatalib.ulb.tu-darmstadt.de/handle/tudatalib/2496.2}{https://tudatalib.ulb.tu-darmstadt.de/handle/tudatalib/2496}} 
    for argument similarity.
    
    We annotate sentential arguments on controversial topics on a continuous scale. We use the dataset by \newcite{Stab2018b}, which contains pro and con stance arguments for eight controversial topics ($T_1$ - $T_8$) (``cloning'', ``abortion'', ``minimum wage'', ``marijuana legalization'', ``nuclear energy'', ``death penalty'', ``gun control'', ``school uniforms'') retrieved from heterogeneous web sources.
 
    Previous work addressing argument similarity \cite{MisraEW16,Reimers:2019:ACL} used discrete scales. However, expressing an inherently continuous property in this way is counter-intuitive and potentially unreliable due to different assumptions made when binning a range of values into a discrete class \cite{RePEc:uwp:landec:v:86:y:2010:iii:1:p530-544}.
   
    Collecting continuous annotations is complex due to selection bias and due to a lack of consistency for a single annotator \cite{kendallcorrelation}. 
    To solve the consistency problem, we apply a comparative approach, which converts the annotation into a preference problem: the annotators stated their preference on pairs of sentential arguments. We utilized the Best-Worst Scaling (\textit{BWS}) method \cite{kiritchenko-mohammad-2016-capturing} to reduce the number of required annotations.
    For each topic regardless of stance, all arguments were randomly paired and for ensuring a certain proportion of similar arguments within the pairings, a distant supervision filtering strategy was implemented by labeling pairs with scores between 0 and 1 using the system proposed by \newcite{MisraEW16}. 
    Next, all argument pairs were sampled with a desired similarity distribution, by creating argument pair bins across three categories: top 1\%, top 2-50\% and remaining pairs. 
    As the final step, we randomly drew pairs from the top 1\% with 50\% probability, and with each 25\% from the two other bins.
    
    The resulting argument pairs were annotated using crowdsourcing via the Amazon Mechanical Turk Platform. 
    For each annotation task, workers were shown four argument pairs and had to select the most and least similar pair amongst them. Each of these tasks was assigned to four different workers. 
    To assess the quality of the resulting annotations, we used split-half reliability measure \cite{10.2307/1434452}.
    Workers' votes were split by half and used to independently rank all argument pairs with the BWS method for each half on each task.
    Finally, the Spearman's rank correlation between the resulting rankings is calculated as a proxy for consistency. 
    The resulting average correlation across all topics in our dataset is 0.66 (random splits are repeated 25 times and final scores averaged), which, given the small number of votes per half (two), is in an acceptable range and reflects the difficulty of this task \cite{kiritchenko-mohammad-2016-capturing}.
    \autoref{tab:bws_topic} lists the mean split-half reliability estimates for all topics (averaged over 25 random splits) in the dataset.
    
    \begin{table}[ht]
    \centering
    \small
    \begin{tabular}{lc|lc}
    \toprule
         \textbf{Topic~$T$} & \textbf{Score} & \textbf{Topic~$T$} & \textbf{Score} \\ \midrule
         Cloning & 0.84 & Nuclear energy & 0.64\\
         Abortion & 0.79 & Death penalty  & 0.58 \\
         Minimum wage & 0.50 & Gun control  & 0.59 \\
         Marijuana legal. & 0.57 & School uniforms  & 0.64 \\
         \midrule
         \multicolumn{4}{c}{Whole dataset = 0.66} \\
        \bottomrule
    \end{tabular}
    \caption{Mean split-half reliability estimate is calculated using Spearman’s rank correlation  $\rho$ per topic $T$ and over the whole \emph{BWS Argument Similarity} dataset.}
    \label{tab:bws_topic}
\end{table}

    We use the resulting \textit{BWS Argument Similarity Dataset} with different splitting strategies in our paper.
    In \emph{cross-topic} tasks, we fix topics ($T_1$ - $T_5$) for training, $T_6$ for development and ($T_7$ and $T_8$) for test sets. This is a difficult task, as models are evaluated on completely unseen topics.

    Note that the cross-topic experiments on this dataset are quite different from cross-domain tasks (\autoref{sec:DomainAdaptationwithAugSBERT}): 
    the model fine-tunes in-domain on fixed topics ($T_1$ - $T_5$ in our case) and is evaluated on unseen topics, whereas in the domain adaptation experiments we fine-tune on target domain data.
    For \emph{in-topic}, we randomly sample fixed and disjoint pairs from each and every topic ($T_1$ - $T_8$) and create our train, development and test splits with approximately equal number of pairs from each topic.

    \textbf{Quora Question Pairs (Quora-QP):} Duplicate question classification identifies whether two questions are duplicates. Quora released a dataset\footnote{\href{{https://www.quora.com/q/quoradata/First-Quora-Dataset-Release-Question-Pairs}}{https://www.quora.com/q/quoradata/First-Quora-Dataset-Release-Question-Pairs}} containing 404,290 question pairs. We start with the same dataset partitions from \newcite{ijcai2017-579}\footnote{ \href{https://drive.google.com/file/d/0B0PlTAo--BnaQWlsZl9FZ3l1c28}{https://drive.google.com/file/d/0B0PlTAo--BnaQWlsZl9FZ3l1c28 }}. We remove all overlaps and ensure that a question in one split of the dataset does not appear in any other split to mitigate the transductive classification problem  \cite{10.1007/978-3-642-15880-3_42}. As we observe performance differences between cross- and bi-encoders mainly for small datasets, we randomly downsample the training set to 10,000 pairs while preserving the original balance of non-duplicate to duplicate question pairs.
    
    \textbf{Microsoft Research Paraphrase Corpus (MRPC):} \newcite{dolan-etal-2004-unsupervised} presented a paraphrase identification dataset consisting of sentence pairs automatically extracted from online news sources. Each pair was manually annotated by two human judges whether they describe the same news event. 
    We use the originally provided train-test splits\footnote{\href{https://github.com/wasiahmad/paraphrase_identification}{https://github.com/wasiahmad/paraphrase\_identification}}. We ensured that all splits have disjoint sentences.

\subsection{Multi-Domain Datasets}

One of the most prominent sentence pair classification tasks with datasets from multiple domains is \textit{duplicate question detection}.
Since our focus is on pairwise sentence scoring, we model this task as a question vs. question (title/headline) binary classification task.
    
    \textbf{AskUbuntu,  Quora,  Sprint, and SuperUser:}
    We replicate the setup of \newcite{shah-etal-2018-adversarial} for domain adaptation experiments. The AskUbuntu and SuperUser data comes from Stack Exchange, which is a family of technical community support forums. Sprint FAQ is a crawled dataset from the Sprint technical forum website. We exclude Apple and Android datasets due to unavailability of labeled question pairs. The Quora dataset (originally derived from the Quora website) is artificially balanced by removing negative question pairs. The statistics for the datasets can be found in \autoref{tab:multi-domain}. 
    Since negative question pairs are not explicitly labeled, \newcite{shah-etal-2018-adversarial} add 100 randomly sampled (presumably) negative question pairs per duplicate question for all datasets except Quora, which has explicit negatives. 
    
\begin{table}[t]
    \centering
    \small
    \resizebox{.49\textwidth}{!}{%
    \begin{tabular}{l c  c  c  c}
    \toprule
    \textbf{Dataset $k$} &  \textbf{Train / Dev / Test} & \textbf{Train} & \textbf{Dev / Test} \\ 
    \textbf{} &  (Total Pairs) & (Ratio) & (Ratio) \\
    \midrule
    \textbf{AskUbuntu} & 919706 / 101k / 101k & 1 : 100 & 1 : 100 \\ 
    \textbf{Quora} & 254142 / 10k / 10k & 3.71 : 100 & 1 : 1\\ 
    \textbf{Sprint} & 919100 / 101k / 101k & 1 : 100 & 1 : 100\\  
    \textbf{SuperUser} & 919706 / 101k / 101k & 1 : 100 & 1 : 100\\ \bottomrule
    \end{tabular}
    }
    \centering
    \caption{Summary of multi-domain datasets originally proposed by \newcite{shah-etal-2018-adversarial} and used for our domain adaptation experiments. Ratio denotes the duplicate pairs (positives) vs. not duplicate pairs (negatives).}
    \label{tab:multi-domain}
\end{table}
  
\begin{table*}[t]
    \centering
    \resizebox{\textwidth}{!}{%
    \small
    \begin{tabular}{l | c | c | c | c  c | c }
         \toprule
         \multicolumn{1}{l}{Task} & \multicolumn{1}{l}{} &
        \multicolumn{3}{c}{Regression ($\rho \times 100$)}    &
        \multicolumn{2}{c}{Classification ($F_1$)}    \\ 
        \cmidrule(lr){3-5}
        \cmidrule(lr){6-7}
        \multicolumn{1}{l}{Model / Dataset} &
        \multicolumn{1}{c}{(Seed Opt.)} & 
        \multicolumn{1}{c}{\textbf{Spanish-STS}} &
        \multicolumn{1}{c}{\textbf{BWS (cross-topic)}} &
        \multicolumn{1}{c}{\textbf{BWS (in-topic)}} &
        \multicolumn{1}{c}{\textbf{Quora-QP}} &
        \multicolumn{1}{c}{\textbf{MRPC}} \\
        \midrule
            Baseline & - & 30.27 & 5.53 & 6.98 & 66.67 & 80.80 \\
            USE \cite{yang2019multilingual} & - & 86.86 & 53.43 & 57.23 & 74.16 & 81.51 \\
        \midrule
            BERT & \xmark & 77.50 $\pm$ 1.49 & 65.06 $\pm$ 1.06  &  65.91 $\pm$ 1.20 & 80.40 $\pm$ 1.05 & 88.95 $\pm$ 0.67  \\
            SBERT & \xmark & 68.36 $\pm$ 5.28 & 58.04 $\pm$ 1.46 & 61.20 $\pm$ 1.66 & 73.44 $\pm$ 0.65 & 84.44 $\pm$ 0.68 \\ 
        \midrule
            BERT (\emph{Upper-bound}) & \cmark &  77.74 $\pm$ 1.24 & 65.78 $\pm$ 0.78 & 66.54 $\pm$ 0.94 & 81.23 $\pm$ 0.93 &  89.00 $\pm$ 0.56 \\
            SBERT (\emph{Lower-bound}) & \cmark & 72.07 $\pm$ 2.05 & 60.54 $\pm$ 0.99 & 63.77 $\pm$ 2.29 & 74.66	 $\pm$ 0.31 & 84.39 $\pm$ 0.51 \\
            SBERT-NLPAug & \cmark & 74.11 $\pm$ 2.58 & 58.15 $\pm$ 1.66 & 61.15 $\pm$ 0.86 & 73.08 $\pm$ 0.42 & 84.47 $\pm$ 0.79 \\            
            
        \midrule

            AugSBERT-R.S. & \cmark & 62.05 $\pm$ 2.53 & 59.95 $\pm$ 0.70 & 64.54 $\pm$ 1.90 & 73.42 $\pm$ 0.74 & 82.28 $\pm$ 0.38 \\
            AugSBERT-KDE & \cmark & 74.67 $\pm$ 1.01 & \textbf{61.49 $\pm$ 0.71} & \textbf{69.76 $\pm$ 0.50} & \textbf{79.31 $\pm$ 0.46} & 84.33 $\pm$ 0.27 \\
            AugSBERT-BM25 & \cmark & 75.08 $\pm$ 1.94 & 61.48 $\pm$ 0.73 & 68.63 $\pm$ 0.79 & 79.01 $\pm$ 0.45 & \textbf{85.46 $\pm$ 0.52} \\
            AugSBERT-S.S. & \cmark & 74.99 $\pm$ 2.30 & 61.05 $\pm$ 1.02 & 68.06 $\pm$ 0.93 & 77.20 $\pm$ 0.41 & 82.42 $\pm$ 0.32 \\
            AugSBERT-BM25+S.S. & \cmark & \textbf{76.24 $\pm$ 1.42} & 59.41 $\pm$ 0.98 & 63.30 $\pm$ 1.34 & 72.45 $\pm$ 0.77 & 82.68 $\pm$ 0.33 \\
        \bottomrule
    \end{tabular}
    }
    \centering
    \caption{Summary of all the datasets being used for the in-domain tasks in this paper. STS and BWS are regression tasks, where we report Spearman's rank correlation $\rho \times 100$. Quora-QP and MRPC are classification tasks, where we report $F_1$ score of the positive class. Scores with the best AugSBERT strategy are highlighted. Corresponding development set performances can be found in \autoref{sec:dev-performances}, \autoref{tab:dev-results}.}
    \label{tab:results}
\end{table*}

\section{Experimental Setup}\label{sec:experimental_setup}
We conduct our experiments using PyTorch Huggingface's transformers \cite{wolf2019huggingfaces} and the sentence-transformers framework\footnote{\href{https://github.com/UKPLab/sentence-transformers}{https://github.com/UKPLab/sentence-transformers}} \cite{reimers-2019-sentence-bert}. The latter showed that BERT outperforms other transformer-like networks when used as bi-encoder. For English datasets, we use \textit{bert-base-uncased}  and for the Spanish dataset we use \textit{bert-base-multilingual-cased}. Every AugSBERT model exhibits computational speeds identical to the SBERT model \cite{reimers-2019-sentence-bert}.
    
    \textbf{Cross-encoders} \quad We fine-tune the BERT-uncased model by optimizing a variety of hyperparameters: hidden-layer sizes, learning-rates and batch-sizes. We add a linear layer with sigmoid activation on top of the \textit{[CLS]} token to output scores 0 to 1. We achieve optimal results with the combination: learning rate of $1\times10^{-5}$, hidden-layer sizes in $\{200,400\}$ and a batch-size of $16$. Refer to  \autoref{tab:BERT-hyper-opt} in \autoref{sec:hyperparameter-tuning}.
  
    \textbf{Bi-encoders} \quad We fine-tune SBERT with a batch-size of $16$, a fixed learning rate of $2\times10^{-5}$, and AdamW optimizer. \autoref{tab:SBERT-hyper-opt} in \autoref{sec:hyperparameter-tuning} lists hyper-parameters we initially evaluated.
    
    \textbf{BM25 and Semantic Search} \quad We evaluate for various top $k$ in $\{3, ..., 18\}$. We conclude the impact of $k$ is not big and overall accomplish best results with $k=3$ or $k=5$ for our experiments. More details in \autoref{sec:top-k}.
    
    \textbf{Evaluation} \quad If not otherwise stated, we repeat our in-domain experiments with 10 different random seeds and report mean scores along with standard deviation. For in-domain regression tasks (STS and BWS), we report the Spearman's rank correlation ($\rho \times 100$) between predicted and gold similarity scores and for in-domain classification tasks (Quora-QP, MRPC), we determine the optimal threshold from the development set and use it for the test set. We report the $F_1$ score of the positive label. For all domain adaptation tasks, we weakly-label the target domain training dataset and measure AUC(0.05) as the metric since it is more robust against false negatives \cite{shah-etal-2018-adversarial}. AUC(0.05) is the area under the curve of the true positive rate as function of the false positive rate (\emph{fpr}), from \emph{fpr} = 0 to \emph{fpr} = 0.05.
    
    \textbf{Baselines} \quad For the in-domain regression tasks, we use  Jaccard similarity to measure the word overlap of the two input sentences. For the in-domain classification tasks, we use a majority label baseline. Further, we compare our results against Universal Sentence Encoder (USE) \cite{yang2019multilingual}, which is a popular state-of-the-art sentence embedding model trained on a wide rang of training data. We utilise the multilingual model\footnote{\href{https://tfhub.dev/google/universal-sentence-encoder-multilingual-large/3}{https://tfhub.dev/google/universal-sentence-encoder-multilingual-large/3}}. Fine-tuning code for USE is not available, hence, we utilise USE as a comparison to a large scale, pre-trained sentence embedding method. Further, we compare our data augmentation strategy AugSBERT against a straightforward data augmentation strategy provided by NLPAug, which implements 15 methods for text data augmentation.\footnote{\href{https://github.com/makcedward/nlpaug}{https://github.com/makcedward/nlpaug}} We include synonym replacement replacing words in sentences with synonyms utilizing a BERT language model. We empirically found synonym-replacement to work best from the rest of the methods provided in NLPAug.
    
\section{Results and Discussion}

\begin{table*}[t]
    \centering
    \small
    \begin{tabular}{l | l |c  c | c | c | c  }
        \toprule
        \multicolumn{2}{l}{} &
        \multicolumn{1}{c}{In-Domain}    &
        \multicolumn{4}{c}{Cross-Domain}    \\ 
        \cmidrule(lr){3-3}
        \cmidrule(lr){4-7}
        \multicolumn{1}{l}{\textbf{Source}} &
        \multicolumn{1}{l}{\textbf{Target}} &
        \multicolumn{1}{c}{\textbf{SBERT}} &
        \multicolumn{1}{c}{\textbf{AugSBERT}} &
        \multicolumn{1}{c}{\textbf{SBERT}} &
        \multicolumn{1}{c}{\textbf{Bi-LSTM}} &
        \multicolumn{1}{c}{\textbf{Bi-LSTM}} \\
        \multicolumn{1}{l}{(Train)} &
        \multicolumn{1}{l}{(Evaluate)} &
        \multicolumn{1}{c}{(\emph{Upper-bound})} &
        \multicolumn{1}{c}{} &
        \multicolumn{1}{c}{(\emph{Lower-bound})} &
        \multicolumn{1}{c}{(Direct)} &
        \multicolumn{1}{c}{(Adversarial)} \\
        \midrule
             & Quora & 0.504 &\textbf{0.496} & \textbf{0.496} & 0.059 & 0.066 \\
            AskUbuntu & Sprint & 0.869 & 0.852 & 0.747 & \textbf{0.93} & 0.923  \\
             & SuperUser & 0.802 & 0.779 &  0.738 & \textbf{0.806} & 0.798   \\
        \midrule
             & AskUbuntu & 0.715  & \textbf{0.602} & 0.501 & 0.351 & 0.328   \\
            Quora & Sprint & 0.869 & \textbf{0.875} & 0.505 & \textbf{0.875} & 0.867    \\
             & SuperUser & 0.802 & \textbf{0.645} & 0.504 & 0.523 & 0.485   \\
        \midrule
             & AskUbuntu & 0.715 & \textbf{0.709} & 0.637 & 0.629 & 0.627  \\
            SuperUser & Quora & 0.504 &\textbf{ 0.495} &  0.495 & 0.058 & 0.067   \\
             & Sprint & 0.869 & 0.876 &  0.785 & 0.936 & \textbf{0.937}   \\
        \midrule
             & AskUbuntu & 0.715 & \textbf{0.663} & 0.613 & 0.519 & 0.543   \\
            Sprint & Quora & 0.504 & 0.495 & \textbf{0.496} & 0.048 & 0.063    \\
             & SuperUser & 0.802 & \textbf{0.769} & 0.660 & 0.658 & 0.636   \\
        \bottomrule
    \end{tabular}
    \centering
    \caption{AUC(0.05) scores for domain adaptation experiments. All except SBERT (in-domain) are evaluated in cross-domain setup with the best transfer strategy highlighted. We adapt \cite{shah-etal-2018-adversarial} Bi-LSTM models. Corresponding development set performances can be found in \autoref{sec:dev-performances}, \autoref{tab:dev-cross-domain}.}
    \label{tab:results-cross-domain}
\end{table*}

\subsection{In-Domain Experiments for AugSBERT}

\autoref{tab:results} summarizes all results for all in-domain datasets. The plain bi-encoder (SBERT w/o Seed Opt.) consistently underperforms (4.5 - 9.1 points) the cross-encoder across all in-domain tasks. Optimizing the seed helps SBERT more than BERT, however, the performance gap remains open (2.8 - 8.2 points). Training with multiple random seeds and selecting the best performing model on the development set can significantly improve the performance. For the smallest dataset (STS), we observe large performance differences between different random seeds. The best and worst seed for SBERT have a performance difference of more than 21 points. For larger datasets, the dependence on the random seed decreases. 
We observe bad training runs can often be identified and stopped early using the early stopping algorithm \cite{dodge2020fine}. 
Detailed results with seed optimization can be found in \autoref{sec:seed-opt}.

Our proposed AugSBERT approach improves the performance for all tasks by 1 up to 6 points, significantly outperforming the existing bi-encoder SBERT and reducing the performance difference to the cross-encoder BERT. It outperforms the synonym replacement data augmentation technique (\textit{NLPAug}) for all tasks. Simple word replacement strategies as shown are not helpful for data augmentation in sentence-pair tasks, even leading to worse performances compared to models without augmentation for BWS and Quora-QP. Compared to the off the shelf USE model, we see a significant improvement with AugSBERT for all tasks except Spanish-STS. This is presumably due to the fact that USE was trained on the SNLI corpus \cite{snli:emnlp2015}, which was used as basis for the Spanish STS test set, i.e., USE has seen the test sentence pairs during training.

For the novel BWS argument similarity dataset, we observe AugSBERT only gives a minor improvement for cross-topic split. We assume this is due to cross-topic setting being a challenging task, mapping sentences of an unseen topic to a vector space such that similar arguments are close. However, on known topics (in-topic),  AugSBERT shows its full capabilities and even outperforms the cross-encoder. 
We think this is due a better generalization of SBERT bi-enconder compared to the BERT cross-encoder. Sentences from known topics (in-topic) are mapped well within a vector space by a bi-encoder. 

\begin{figure}[htb!]
    \begin{center}
        \includegraphics[width=1.0\linewidth]{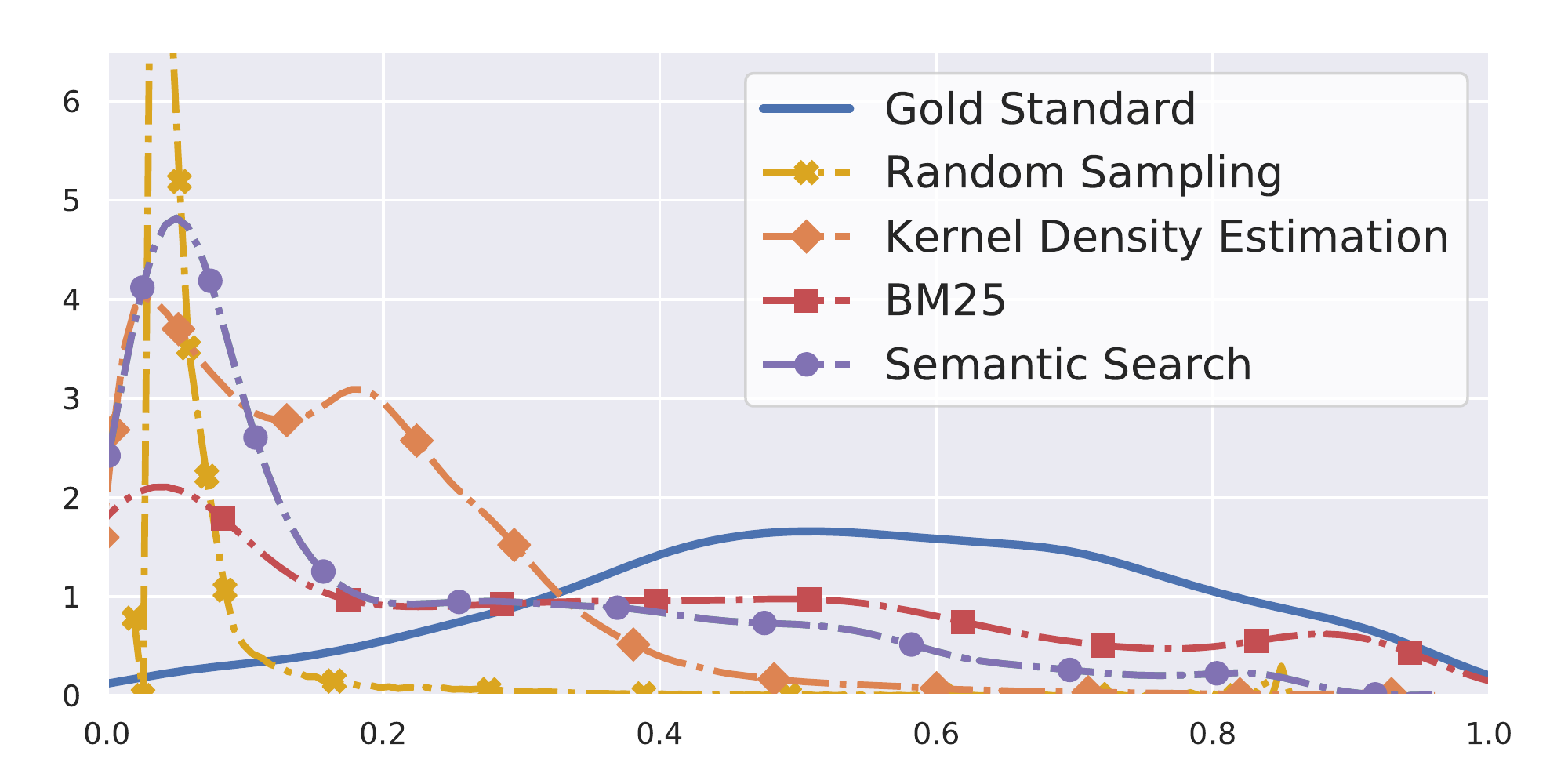}
        \captionof{figure}{Comparison of the density distributions of gold standard with silver standard for various sampling techniques on Spanish-STS (in-domain) dataset.}
        \label{tab:distributions}
    \end{center}
\end{figure}
    
\textbf{Pairwise Sampling} We observe that the sampling strategy is critical to achieve an improvement using AugSBERT. Random sampling (R.S.) decreases performance compared to training SBERT without any additional silver data in most cases. 
BM25 sampling and KDE produces the best AugSBERT results, followed by Semantic Search (S.S.). 
\autoref{tab:distributions}, which shows the score distribution for the gold and silver dataset for Spanish-STS, visualizes the reason for this. 
With random sampling, we observe an extremely high number of low similarity pairs. This is expected, as randomly sampling two sentences yields in nearly all cases a dissimilar pair.  
In contrast, \textit{BM25} generates a silver dataset with similar score distribution to the gold training set. 
It is still skewed towards low similarity pairs, but has the highest percentage of high similarity pairs. 
\textit{BM25+S.S.}, apart on Spanish-STS, overall performs worse in this combination than the individual methods. It even underperforms random sampling on the BWS and Quora-QP datasets. We believe this is due to the aggregation of a high number of dissimilar pairs from the sampling strategies combined.
\textit{KDE} shows the highest performance in three tasks, but only marginally outperforms BM25 in two of these. 
Given that BM25 is the most computationally efficient sampling strategy and also creates smaller silver datasets (numbers are given in \autoref{sec:silver-sizes}, \autoref{tab:silver}), it is likely the best choice for practical applications.

\subsection{Domain Adaptation with AugSBERT}
We evaluate the suitability of AugSBERT for the task of domain adaptation. We use duplicate question detection data from different (specialized) online communities. Results are shown in
\autoref{tab:results-cross-domain}. We can see in almost all combinations that AugSBERT outperforms SBERT trained on out-of-domain data (cross-domain). 
On the Sprint dataset (target), the improvement can be as large as 37 points. In few cases, AugSBERT even outperforms SBERT trained on gold in-domain target data.

We observe that AugSBERT benefits a lot when the source domain is rather generic (e.g.\ Quora) and the target domain is rather specific (e.g.\ Sprint).
We assume this is due to Quora forum covering many different topics including both technical and non-technical questions, transferred well by a cross-encoder to label the specific target domain (thus benefiting AugSBERT). Vice-versa, when we go from a specific domain (Sprint) to a generic target domain (Quora), only a slight performance increase is noted.

For comparison, \autoref{tab:results-cross-domain} also shows the state-of-the-art results from \newcite{shah-etal-2018-adversarial}, who applied direct and adversarial domain adaptation with a Bi-LSTM bi-encoder. With the exception of the Sprint dataset (target), we outperform that system with substantial improvement for many combinations.

\section{Conclusion}

We presented a simple, yet effective data augmentation approach called AugSBERT to improve bi-encoders for pairwise sentence scoring tasks. The idea is based on using a more powerful cross-encoder to soft-label new sentence pairs and to include these into the training set. 

We saw a performance improvement of up to 6 points for in-domain experiments. However, selecting the right sentence pairs for soft-labeling is crucial and the naive approach of randomly selecting pairs fails to achieve a performance gain. We compared several sampling strategies and found that BM25 sampling provides the best trade-off between performance gain and computational complexity. 

The presented AugSBERT approach can also be used for domain adaptation, by soft-labeling data on the target domain. In that case, we observe an improvement of up to 37 points compared to an SBERT model purely trained on the source domain.

\section*{Acknowledgements}

This work has been supported by the German Federal Ministry of Education and Research (BMBF) under the promotional reference 03VP02540 (ArgumenText), by the German Research Foundation through the German-Israeli Project Cooperation (DIP, grant DA 1600/1-1 and
grant GU 798/17-1) and has been funded by the German Federal Ministry of Education and Research and the Hessian Ministry of Higher Education, Research, Science and the Arts within their joint support of the National Research Center for Applied Cybersecurity ATHENE. We would like to thank Andreas Rücklé, Jan-Christoph Klie, Mohsen Mesgar, Kevin Stowe and the anonymous reviewers for their feedback.

\bibliography{paper}
\bibliographystyle{acl_natbib}

\appendix
\section{Appendices}
\label{sec:appendix}

In this appendix, we mention the following sections in detail: MTurk guidelines and density distribution analysis for the BWS argument similarity dataset~(\ref{sec:bws}), hyperparameter-tuning (\ref{sec:hyperparameter-tuning}) and seed-optimization (\ref{sec:seed-opt});
provide analysis of the top-$k$ parameter (\ref{sec:top-k}) and computational efficiency (\ref{sec:silver-sizes}) for our in-domain sampling strategies; report development set performances for all our tasks (\ref{sec:dev-performances}).
 
\section{BWS Argument Similarity Dataset}\label{sec:bws}

\subsection{Amazon Mechanical Turk Guidelines}
The annotations required for the BWS Argument Similarity Corpus were acquired via crowdsourcing on the
Amazon Mechanical Turk platform. Workers participating in the study had to be located in the US,
with more than 100 HITs approved and an overall
acceptance rate of 90\% or higher. 
We paid them at the US federal minimum wage of \$7.25/hour.
Workers also had to qualify for the study by passing a qualification test consisting of four test
questions with argument pairs. \autoref{fig:amt-guidelines} exemplifies the instructions given to workers.

\subsection{Density Distribution Analysis}

\begin{figure}[t]
    \begin{center}
        \includegraphics[width=0.95\linewidth]{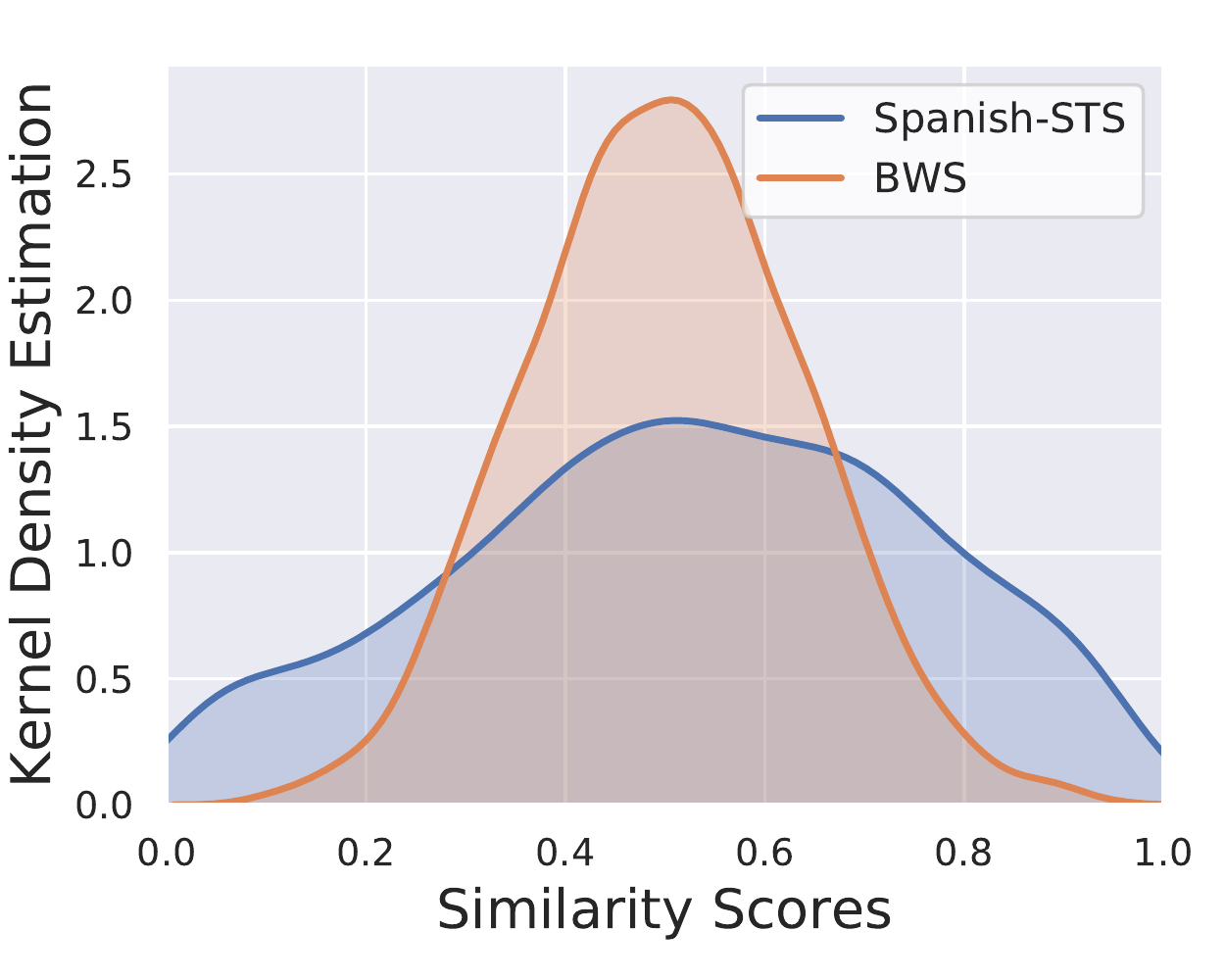}
        \captionof{figure}{Comparison of density distribution of BWS Argument Similarity dataset with Spanish-STS dataset.}
        \label{fig:bws-vs-sts}
        \vspace{-3.5mm}
    \end{center}
    
\end{figure}

\autoref{fig:bws-vs-sts}~compares the density distributions of BWS with Spanish-STS. 
For the Spanish-STS dataset, the pre-sampling process results in a high amount of pairs towards either ends of the similarity scale---leading to selection bias. 
The pre-sampling of the creation process of the BWS dataset, in turn, is less biased.
There is a much lower number of pairs towards either end of the scale, which is in accordance with data from the wild, i.e. randomly paired arguments. 

\section{Hyperparameter Tuning} \label{sec:hyperparameter-tuning}
    We implement \textit{coarse to fine} random search to find the optimal combination of hyperparameters for both cross-encoders (BERT) and bi-encoders (SBERT). We choose the optimal combination based on the development dataset performance keeping random seed value fixed.\footnote{Random seed value = 42 during hyperparameter tuning experiments.} 
    
    \paragraph{Cross-Encoder (BERT):} For all fine-tuning experiments, we optimize a variety of hyperparameters: hidden-layer sizes, learning-rates and batch-sizes. We first evaluate over a wide range of parameters and later conduct a deeper fine search of these optimal parameters. Experimental setup can be found in \autoref{tab:BERT-hyper-opt}.
    
    \begin{table}[h!]
    \centering
    \small
        \begin{tabular}{c|c }
         \toprule
            BERT model & bert-base (uncased/multi.-cased) \\ \midrule
            hidden layer sizes & \{100, 200, 400, 800, 1600, 3200\} \\ \midrule
            Learning rates & \{1e-4, 1e-5, 1e-6\} \\ \midrule
            Batch sizes & \{8, 16\}  \\ 
         \bottomrule
        \end{tabular}
    \caption{Experimental setup for hyperparameter tuning of cross-encoder (BERT).}
    \label{tab:BERT-hyper-opt}
    \end{table}
    
    \paragraph{Bi-Encoder (SBERT):} For all fine-tuning experiments, we utilize \textit{bert-base} models, and implement coarse to fine random search with various learning-rates and batch-sizes. 
    Since changing the learning rate scheduler did not contribute to significant improvement, we kept it constant for all our experiments. Experimental setup can be found in \autoref{tab:SBERT-hyper-opt}.

    \begin{table}[h]
    \centering
    \small
    \resizebox{.48\textwidth}{!}{%
        \begin{tabular}{c|c}
         \toprule
            BERT model & bert-base (uncased/ multi.-cased) \\ \midrule
            Learning rates & \{2e-5, 1e-6, 1e-7\} \\ \midrule
            Learning rate scheduler &  constant \\ \midrule
            Batch sizes & \{8, 16\} \\ 
         \bottomrule
        \end{tabular}
        }
    \caption{Experimental setup for hyperparameter tuning of bi-encoder (SBERT).}
    \label{tab:SBERT-hyper-opt}
    \end{table}

\section{Seed Optimization}\label{sec:seed-opt}

\begin{figure}[htb]
	\centering
	\includegraphics[width=\linewidth]{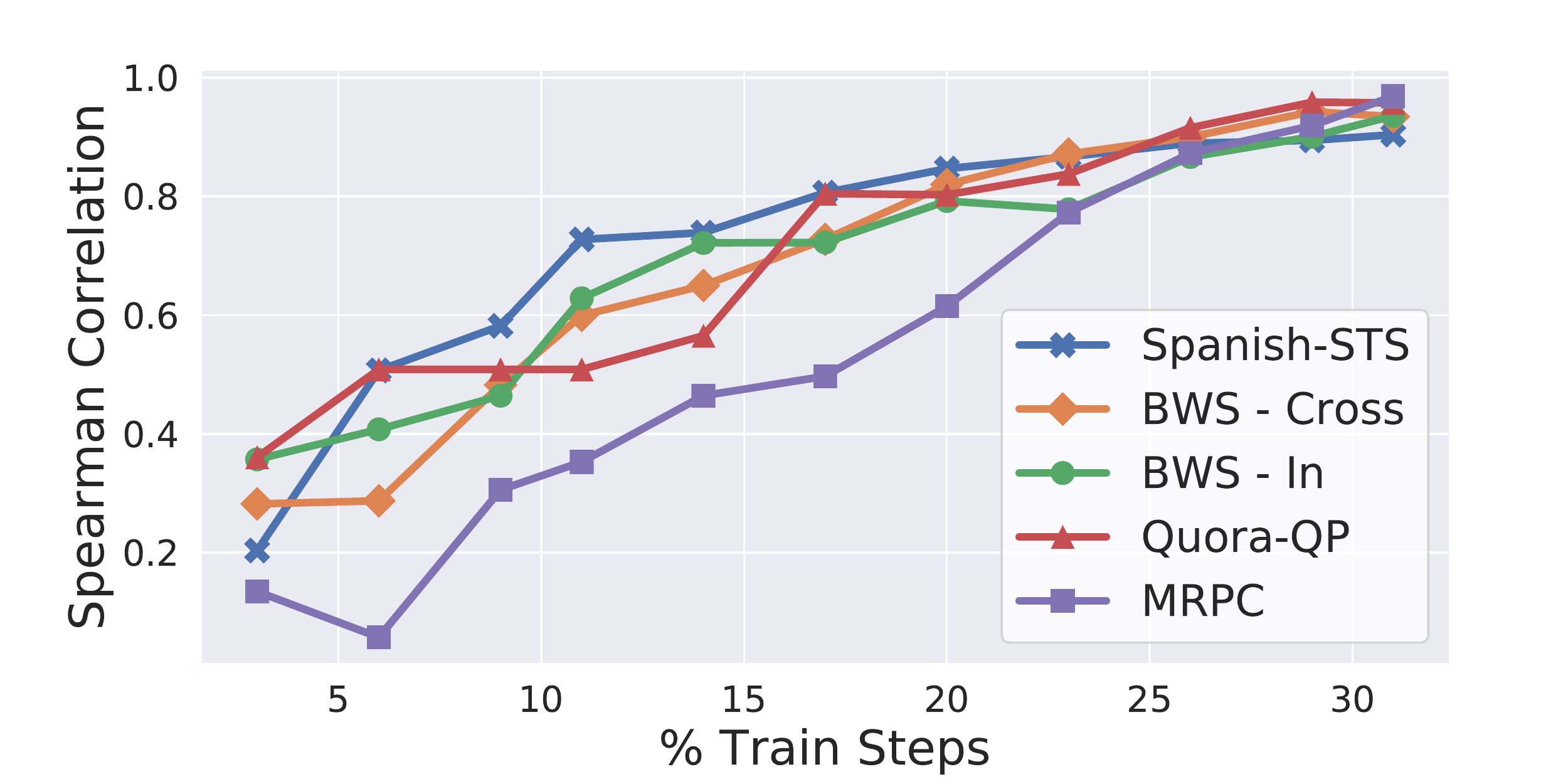}
	\caption{Spearman's rank correlation for SBERT bi-encoder between development scores at x\% of the training steps with final development score for in-domain datasets.}
	\label{fig:dev_correlation}
\end{figure}

 For our in-domain tasks, we apply seed optimization i.e. we train our models with 5 random seeds and select the model that performs best on the development set, and repeat this complete setup 10 times. Testing various seeds can be computationally expensive. In order to reduce the computational overhead, we evaluate whether bad runs can be identified and stopped early. At x\% of the overall training steps we evaluate the model on the development set and compare the rank with the final ranking of the models on the development set. The results are depicted in \autoref{fig:dev_correlation}. We observe a Spearman's rank correlation of over 0.8 at about 20\% of the training steps. We conclude, that bad training runs can often be identified and stopped early.

\section{Impact of Top K in Sampling Strategy}\label{sec:top-k}

    In sampling strategies, such as BM25 and semantic search, we are required to pick the top $k$ values returned by the retrieval engine. Typically for small $k$ values, positive-pairs are dominant and with increase in $k$, negative-pairs start becoming dominant. 
    
    We chose a top $k$ value within \{3,5,7,9,12,18\} and evaluated the final scores retrieved from our experiments, to measure an impact of $k$. 
    Overall, we find the impact of $k$ to be rather small and $k=3$ or $k=5$ producing optimal scores for most of the experiments. Top-$k$ mean test scores for our in-domain datasets are reported in \autoref{tab:BM25} for BM25 and \autoref{tab:Semantic-Search} for semantic search sampling strategies respectively.

\begin{table*}[htb!]
    \centering
    \small
    \begin{tabular}{ l| c | c  c  c  c  c  c}
    \toprule
    Dataset/Top $k$ & Measure & Top 3 & Top 5 & Top 7 & Top 9 & Top 12 & Top 18 \\ \midrule
    Spanish-STS & $\rho \times 100$ & 73.67 & \textbf{75.08} & 74.83 & 74.71 & 73.89 & 72.82 \\ 
    BWS (cross) & $\rho \times 100$ & 60.02 & 60.23 & \textbf{61.48} & 60.65 & 60.89 & 61.47 \\ 
    BWS (in) & $\rho \times 100$ & 66.26 & \textbf{68.63} & 67.49 & 67.38 & 67.74 & 68.08 \\ 
    Quora-QP & $F_1$ & \textbf{79.01} & 78.68 & 78.49 & 78.40 & 78.46 & 77.75 \\ 
    MRPC & $F_1$ & \textbf{85.46} & 85.17 & 85.03 & 84.15 & 84.24 & 84.27 \\ \bottomrule
    \end{tabular}
    \centering
    \caption{Summary of In-domain BM25 Sampling Strategy: Top $k$ mean test scores. We report Spearman's rank correlation $\rho \times 100$ for regression tasks and $F_1$ score for classification tasks.}
    \label{tab:BM25}
\end{table*}

\begin{table*}[htb!]
    \centering
    \small
    \begin{tabular}{l | c | c  c  c  c  c  c}
    \toprule
    Dataset/Top $k$ & Measure & Top 3 & Top 5 & Top 7 & Top 9 & Top 12 & Top 18 \\ \midrule
    Spanish-STS & $\rho \times 100$ & 73.84 & 74.22 & \textbf{74.99} & 74.31 & 74.22 &73.61 \\ 
    BWS (cross) & $\rho \times 100$ & 60.57 & 60.51 & 60.76 & \textbf{61.04} & 60.74 & 60.87 \\ 
    BWS (in) & $\rho \times 100$ & 65.39 & \textbf{68.06} & 67.01 & 66.78 & 66.95 & 65.93 \\ 
    Quora-QP & $F_1$ & \textbf{77.20} & 76.65 & 76.41 & 76.68 & 76.33 & 76.32 \\ 
    MRPC  & $F_1$ & \textbf{82.42} & 82.18 & 82.20 & 81.86 & 81.81 & 81.91 \\ \bottomrule
    \end{tabular}
    \centering
    \caption{Summary of In-domain Semantic Search Sampling Strategy: Top $k$ mean test scores. We report Spearman's rank correlation $\rho \times 100$ for regression tasks and $F_1$ score for classification tasks.}
    \label{tab:Semantic-Search}
\end{table*}

\section{Computational Efficiency vs. Size of Silver Datasets}\label{sec:silver-sizes}
    
    The augmented SBERT strategy requires to weakly label a large set of sentence pairs with the cross-encoder. 
    The larger the set of silver pairs, the bigger is the overhead for labeling with the cross-encoder and subsequent training the bi-encoder. 
    Hence, for reasons of efficiency, smaller silver dataset sizes are preferable. 
    \autoref{tab:silver} summarizes the performance of each sampling technique versus the size of sampled silver pairs. 
    
    Different sampling strategies create vastly different amounts of sentence pairs. Randomly sampling (R.S.) a large number of sentence pairs is not efficient and often leads to worse performances. KDE with large silver datasets produce optimal scores, but is less computationally efficient. 
    Semantic Search (S.S.) requires the bi-encoder to be additionally trained, which causes computational overhead. Finally, BM25 overall on an average performs best for all tasks given computational efficiency, by sampling out the smallest silver dataset sizes for all tasks in \autoref{tab:silver}.
    
\section{Development Set Performances}\label{sec:dev-performances}

    The development set performances for all sentence pair in-domain and domain adaptation tasks can be referred in \autoref{tab:dev-results} and \autoref{tab:dev-cross-domain} respectively.

\begin{table*}[t]
    \centering
    \small
    \resizebox{.98\textwidth}{!}{%
        \begin{tabular}{l| c  | c  c | c  c | c  c | c  c | c  c}
        \toprule
        
        \multicolumn{1}{l}{Sampling Tech.} &
        \multicolumn{1}{c}{\textbf{None}}    &
        \multicolumn{2}{c}{\textbf{BM25}} &
        \multicolumn{2}{c}{\textbf{Sem. Search}} & 
        \multicolumn{2}{c}{\textbf{BM25 + S.S.}} & 
        \multicolumn{2}{c}{\textbf{KDE}} &
        \multicolumn{2}{c}{\textbf{Random Samp.}} \\ 
        \cmidrule(lr){2-2}
        \cmidrule(lr){3-4}
        \cmidrule(lr){5-6}
        \cmidrule(lr){7-8}
        \cmidrule(lr){9-10}
        \cmidrule(lr){11-12}
        \multicolumn{1}{l}{Dataset} &
        \multicolumn{1}{c}{Score} & 
        \multicolumn{1}{c}{(\#Silver)} &
        \multicolumn{1}{c}{Score} & 
        \multicolumn{1}{c}{(\#Silver)} &
        \multicolumn{1}{c}{(Score)} & 
        \multicolumn{1}{c}{\#Silver} &
        \multicolumn{1}{c}{(Score)} & 
        \multicolumn{1}{c}{(\#Silver)} &
        \multicolumn{1}{c}{(Score)} & 
        \multicolumn{1}{c}{(\#Silver)} 
        & \multicolumn{1}{c}{(Score)} \\
        
        
        \midrule
        Spanish-STS & 72.07 & \textbf{3,964} & 75.08 & 12,715 & 74.99 & 16,678 & \textbf{76.24} & 230,364 &  74.67 &  911,072 & 62.05\\ \midrule
        BWS (cross-topic) & 60.54 & \textbf{11,694} & 61.48 & 18,824 & 61.05 & 28,771 & 59.41 & 559,630 & \textbf{61.49} & 72,3540 & 59.95 \\ \midrule
        BWS (in-topic)  & 63.77 & \textbf{9,236} & 68.63 & 11,816 & 68.06 & 19,830 & 63.30 & 394,252 & \textbf{69.76} &  565,820 & 64.54\\ \midrule
        Quora-QP & 74.66 & \textbf{28,014} & 79.01 & 47,055 & 77.20 & 75,067 & 72.45 & 50,147 & \textbf{79.31} &  1,000,000 & 73.42 \\\midrule
        MRPC  & 84.39 & \textbf{10,637} & \textbf{85.46} & 18,292 & 82.39 & 25,867 & 82.68 & 32,353 & 84.33 & 1,000,000 & 82.28 \\ \bottomrule
        \end{tabular}
        }
        \caption{Summary of (\textit{\#silver dataset samples}, \textit{mean score}) for each sampling technique across all in-domain datasets. For STS and BWS datasets, we report the Spearman's rank correlation $\rho \times 100$ and the $F_1$ score for Quora-QP and MRPC datasets. None represents plain bi-encoder i.e. SBERT. Scores with best sampling strategy and smallest silver dataset size across each dataset are highlighted.}
        \label{tab:silver}
    \end{table*}
    
\begin{table*}[t]
    \centering
    \resizebox{\textwidth}{!}{%
    \small
    \begin{tabular}{l | c | c | c | c  c | c }
         \toprule
         \multicolumn{1}{l}{Task} & \multicolumn{1}{l}{} &
        \multicolumn{3}{c}{Regression ($\rho \times 100$)}    &
        \multicolumn{2}{c}{Classification ($F_1$)}    \\ 
        \cmidrule(lr){3-5}
        \cmidrule(lr){6-7}
        \multicolumn{1}{l}{Model / Dataset} &
        \multicolumn{1}{c}{(Seed-Opt.)} & 
        \multicolumn{1}{c}{\textbf{Spanish-STS}} &
        \multicolumn{1}{c}{\textbf{BWS (cross-topic)}} &
        \multicolumn{1}{c}{\textbf{BWS (in-topic)}} &
        \multicolumn{1}{c}{\textbf{Quora-QP}} &
        \multicolumn{1}{c}{\textbf{MRPC}} \\
        \midrule
            Baseline & - & 16.98 & 6.31 & 5.06 & 66.67 & 80.75 \\
        \midrule
            BERT & \xmark & 89.10 $\pm$ 0.69& 60.97 $\pm$ 1.35 & 64.89 $\pm$ 1.41 & 81.87 $\pm$ 1.07 & 89.71 $\pm$ 0.54  \\
            SBERT & \xmark & 82.15 $\pm$ 0.95 & 54.77 $\pm$ 0.68 & 62.66 $\pm$ 1.06 & 76.69 $\pm$ 0.34 & 87.30 $\pm$ 0.45 \\ 
        \midrule
            BERT (\emph{Upper-bound}) & \cmark & 88.86 $\pm$ 0.74 & 62.17 $\pm$ 0.83 & 66.23 $\pm$ 0.96 & 81.64 $\pm$ 0.99 & 89.75 $\pm$ 0.46 \\
            SBERT (\emph{Lower-bound}) & \cmark & 82.30 $\pm$ 1.11 & 54.90 $\pm$ 0.88 & 62.75 $\pm$ 1.16 & 76.73 $\pm$ 0.39 & 87.37 $\pm$ 0.52 \\
            SBERT-NLPAug & \cmark & 84.63 $\pm$ 0.77 & 58.66 $\pm$ 0.49 & 66.16 $\pm$ 0.41 & 79.72 $\pm$ 0.41 & 87.24 $\pm$ 0.52 \\

        \midrule

            AugSBERT-R.S. & \cmark & 75.90 $\pm$ 1.89 & 57.36 $\pm$ 0.80 & 67.83 $\pm$ 0.43 & 73.42 $\pm$ 0.74 & 84.12 $\pm$ 0.64 \\
            AugSBERT-KDE & \cmark & 85.69 $\pm$ 0.54 & 57.71 $\pm$ 0.85 & 66.14 $\pm$ 0.72 & \textbf{80.00 $\pm$ 0.31} & 87.04 $\pm$ 0.29\\
            AugSBERT-BM25 & \cmark  & 85.10 $\pm$ 1.11 & \textbf{57.95 $\pm$ 0.98} & 65.37 $\pm$ 1.18 & 77.73 $\pm$ 0.47 & \textbf{88.04 $\pm$ 0.51} \\
            AugSBERT-S.S. & \cmark & 85.28 $\pm$ 0.85 & 56.84 $\pm$ 1.22 & 66.21 $\pm$ 0.92 & 79.29 $\pm$ 0.31 & 83.52 $\pm$ 0.27 \\
            AugSBERT-BM25+S.S. & \cmark & \textbf{85.98 $\pm$ 0.75} & 57.83 $\pm$ 0.69 & \textbf{68.45 $\pm$ 0.57} & 79.75 $\pm$ 0.24 & 85.26 $\pm$ 0.56 \\
        \bottomrule
    \end{tabular}
    }
    \centering
    \caption{Summary of development set scores for the in-domain tasks in this paper. STS and BWS are regression tasks, where we report Spearman's rank correlation $\rho \times 100$. Quora-QP and MRPC are classification tasks, where we report $F_1$ score of the positive class. For Baselines, we use a simple Jaccard similarity for regression tasks and a majority label baseline for classification tasks. Scores with the best augmented SBERT strategy are highlighted.}
    \label{tab:dev-results}
\end{table*}

\begin{table*}[t]
    \centering
    \small
    \begin{tabular}{l | l |c  c | c | c | c  }
        \toprule
        \multicolumn{2}{l}{} &
        \multicolumn{1}{c}{In-Domain}    &
        \multicolumn{4}{c}{Cross-Domain}    \\ 
        \cmidrule(lr){3-3}
        \cmidrule(lr){4-7}
        \multicolumn{1}{l}{\textbf{Source}} &
        \multicolumn{1}{l}{\textbf{Target}} &
        \multicolumn{1}{c}{\textbf{SBERT}} &
        \multicolumn{1}{c}{\textbf{AugSBERT}} &
        \multicolumn{1}{c}{\textbf{SBERT}} &
        \multicolumn{1}{c}{\textbf{Bi-LSTM}} &
        \multicolumn{1}{c}{\textbf{Bi-LSTM}} \\
        \multicolumn{1}{l}{(Train)} &
        \multicolumn{1}{l}{(Evaluate)} &
        \multicolumn{1}{c}{(\emph{Upper-bound})} &
        \multicolumn{1}{c}{} &
        \multicolumn{1}{c}{(\emph{Lower-bound})} &
        \multicolumn{1}{c}{(Direct)} &
        \multicolumn{1}{c}{(Adversarial)} \\
        \midrule
             & Quora & 0.675  & \textbf{0.638} & 0.496 & 0.062 & 0.071 \\
            AskUbuntu & Sprint & 0.989 & 0.773 & 0.670 & \textbf{0.921} & 0.917  \\
             & SuperUser & 0.908 & \textbf{0.801} & 0.586 & 0.797 & 0.782 \\
        \midrule
             & AskUbuntu & 0.844  & \textbf{0.512} &  0.511 & 0.328 & 0.309   \\
            Quora & Sprint & 0.989 & 0.675 & 0.524 & 0.639 & \textbf{0.848}   \\
             & SuperUser & 0.908 & 0.509  &  0.512 & \textbf{0.529} & 0.473   \\
        \midrule
             & AskUbuntu & 0.844 & 0.612 &  0.564 & 0.607 & \textbf{0.620}  \\
            SuperUser & Quora & 0.675 & \textbf{0.672} & 0.496 & 0.066 & 0.077   \\
             & Sprint & 0.989 & \textbf{0.848} &  0.650 & 0.936 & 0.933   \\
        \midrule
             & AskUbuntu & 0.844 & \textbf{0.724} & 0.601 & 0.521 & 0.532   \\
            Sprint & Quora & 0.511 & \textbf{0.668} & 0.497 & 0.049 & 0.063    \\
             & SuperUser & 0.908 & \textbf{0.748} & 0.620 & 0.652 & 0.631   \\
        \bottomrule
    \end{tabular}
    \centering
    \caption{AUC(0.05) development scores for domain adaptation experiments. All except SBERT (in-domain) are evaluated in cross-domain setup with the best transfer strategy highlighted. We adapt \cite{shah-etal-2018-adversarial} Bi-LSTM models.}
    \label{tab:dev-cross-domain}
\end{table*}

\begin{figure*}[t]
\fbox{%
\small{
	\begin{minipage}{\textwidth}
        Arguments are \textbf{\textit{similar}} if -
        
        \begin{itemize}
		    \item They say exactly the same thing in different words. Example for topic ``Fracking'',
		    
		    Argument A: ``\textit{And the toxic chemicals associated with fracking operations can contaminate the soil, air and water, and leach into crops}''.
		
		 	Argument B: ``\textit{The chemicals used in fracking are toxic and threaten to poison and pollute our air, ground, water and food supplies - basic necessities for life}''. 
		 	
		    \item They cover the same aspect and only differ in minor details. Example for topic ``Electric Cars'',
		    
		    Argument A: ``\textit{With literally hundreds of moving parts, a petro-fired automobile requires considerably more maintenance than an electric car}''.
		
		 	Argument B: ``\textit{Electric cars are much more reliable and require less maintenance than gas-powered cars}''.
		    
		    \item They talk about the same general aspect but differ in important details. Example for topic ``Electric Cars'',

		    Argument A: ``\textit{Electric cars are environmentally friendly as it reduces air pollution}''.
		
		 	Argument B: ``\textit{Many people think that electric cars are better than gasoline models, not only because of lower operating costs, but because of quicker acceleration and cleaner air}''.

		\end{itemize}
		Arguments are \textbf{\textit{not similar}} if
		\begin{itemize}
		    \item They have the same topic but do not cover the same aspect. Example for topic ``Electric cars'',
		    
		    Argument A: ``\textit{Electric cars are environmentally friendly as it reduces air pollution}''.
		
		 	Argument B: ``\textit{Generally electric motors for automobiles are much easier to maintain}''.	
		    
		    \item They have different topics. Example for topic ``Robotic Surgery'',
		    
		    Argument A: ``\textit{Opponents argue that more drilling offshore could damage sensitive ecosystems}''.
		
		 	Argument B: ``\textit{Robotic surgery offers patients less pain, fewer complications, and a faster return to normal daily activities}''.	
		\end{itemize}
\end{minipage}
}}
    \caption{Amazon Mechanical Turk HIT Guidelines used in the annotation study for the BWS Argument Similarity Corpus.}
    \label{fig:amt-guidelines}
\end{figure*}

\end{document}